\documentclass[12pt,lettersize,journal,onecolumn]{IEEEtran}
\usepackage{algorithmic}
\usepackage{array}
\usepackage[caption=false,font=normalsize,labelfont=sf,textfont=sf]{subfig}
\usepackage{textcomp}
\usepackage{stfloats}
\usepackage{url}
\usepackage{verbatim}
\usepackage{graphicx}
\usepackage{amsmath,amssymb,amsfonts}%
\usepackage{cite}

\usepackage[utf8]{inputenc}
\usepackage[T1]{fontenc}

\usepackage{multirow}%
\usepackage{amsthm}%
\usepackage{mathrsfs}%
\usepackage{xcolor}%
\usepackage{textcomp}%
\usepackage{manyfoot}%
\usepackage{booktabs}%
\usepackage{algorithm}%
\usepackage{listings}%
\usepackage{hyperref}
\usepackage{setspace}
\usepackage{caption}
\def\BibTeX{{\rm B\kern-.05em{\sc i\kern-.025em b}\kern-.08em
    T\kern-.1667em\lower.7ex\hbox{E}\kern-.125emX}}
\usepackage{balance}
\usepackage{setspace}
\begin{document}
\title{Multi-modal Iterative and Deep Fusion Frameworks for Enhanced Passive DOA Sensing via a Green Massive H$^2$AD MIMO Receiver}
\author{Jiatong Bai, Minghao Chen, Wankai Tang, Yifan Li, Cunhua Pan, Yongpeng Wu, Feng Shu
\thanks{This work was supported in part by the National Natural Science Foundation of China under Grant U22A2002, and  by the Hainan Province Science and Technology Special Fund under Grant ZDYF2024GXJS292; in part by the Scientific Research Fund Project of Hainan University under Grant KYQD(ZR)-21008; in part by the Collaborative Innovation Center of Information Technology, Hainan University, under Grant XTCX2022XXC07; in part by the National Key Research and Development Program of China under Grant 2023YFF0612900. (Corresponding author: Feng Shu).}

\thanks{Jiatong Bai, Minghao Chen is with the School of Information and Communication Engineering, Hainan University, Haikou, 570228, China. (e-mail: 18419229733@163.com; chenminghao2023@126.com).}

\thanks{Wankai Tang is with the National Mobile Communications Research Laboratory, Southeast University, Nanjing, 210096, China. (e-mail: tangwk@seu.edu.cn).}

\thanks{Yifan Li is with the School of Electronic and Optical Engineering, Nanjing University of Science and Technology, Nanjing 210094, China. (e-mail: liyifan97@foxmail.com).}

\thanks{Cunhua Pan is with the National Mobile Communications Research Laboratory, Southeast University, Nanjing, China. (e-mail: c.pan@qmul.ac.uk).}

\thanks{Yongpeng Wu is with the Department of Electronic Engineering, Shanghai Jiao Tong University, Shanghai, 200240, China. (e-mail: yongpeng.wu@sjtu.edu.cn).}

\thanks{Feng Shu is with the School of Information and Communication Engineering and Collaborative Innovation Center of Information Technology, Hainan University, Haikou 570228, China, and also with the School of Electronic and Optical Engineering, Nanjing University of Science and Technology, Nanjing 210094, China. (e-mail: shufeng0101@163.com).}
}

\markboth{Journal of \LaTeX\ Class Files,~Vol.~14, No.~8, August~2021}%
{Shell \MakeLowercase{\textit{et al.}}: A Sample Article Using IEEEtran.cls for IEEE Journals}

\IEEEpubid{0000--0000/00\$00.00~\copyright~2021 IEEE}

\maketitle

\begin{abstract}
Most existing DOA estimation methods assume ideal source incident angles with minimal noise. Moreover, directly using pre-estimated angles to calculate weighted coefficients can lead to performance loss. Thus, a green multi-modal (MM) fusion DOA framework is proposed to realize a more practical, low-cost and high time-efficiency DOA estimation for a H$^2$AD array. Firstly, two more efficient clustering methods, global maximum cos\_similarity clustering (GMaxCS) and global minimum distance clustering (GMinD), are presented to infer more precise true solutions from the candidate solution sets. Based on this, an iteration weighted fusion (IWF)-based method is introduced to iteratively update weighted fusion coefficients and the clustering center of the true solution classes by using the estimated values. Particularly, the coarse DOA calculated by fully digital (FD) subarray, serves as the initial cluster center.
The above process yields two methods called MM-IWF-GMaxCS and MM-IWF-GMinD. To further provide a higher-accuracy DOA estimation, a fusion network (fusionNet) is proposed to aggregate the inferred two-part true angles and thus generates two effective approaches called 
MM-fusionNet-GMaxCS and MM-fusionNet-GMinD.
The simulation outcomes show the proposed four approaches can achieve the ideal DOA performance and the CRLB. Meanwhile, proposed MM-fusionNet-GMaxCS and MM-fusionNet-GMinD exhibit superior DOA performance compared to MM-IWF-GMaxCS and MM-IWF-GMinD, especially in extremely-low SNR range.
\end{abstract}


\section{Introduction}
\IEEEPARstart{W}{ith} the advancement of fifth generation (5G) and 6G mobile communication technologies such as massive multiple-input multiple-output (MIMO), the research on the combination of massive MIMO and direction of arrival (DOA) estimation \cite{shuHADDOA2018tcom, 11111,10175652} has gradually emerged as a hot spot. And it is widely used in target orientation identification and tracking\cite{9966507}, satellite detection\cite{10083213}, radar\cite{9759488}, directional modulation\cite{zhuang2020machine}, as well as green wireless communications and networks\cite{9933838,9687669}, etc.

\subsection{Traditional-based methods}
Regular beamforming\cite{Bartlett1992PropertiesOS,9954622}, as one of the traditional DOA estimation methods, applied the Fourier transform to spatial spectrum estimation, 
but its angular resolution was affected by the “Rayleigh resolution limit”. To alleviate this limitation, the minimum variance method based on the least-mean-square criterion was proposed to improve the spatial resolving power\cite{1449208}. However, achieving high-precision estimation with it required higher signal-to-noise ratios (SNR) and a greater number of snapshots.
Subsequently, the multiple signal classification (MUSIC) method \cite{schmidt1982signal}, the rotation-invariant subspace (ESPRIT) method \cite{roy1986esprit}, and various efficient derivative algorithms \cite{rao1989performance}, such as the root-MUSIC, were proposed. However, these subspace classification methods can cause significant degradation of DOA performance due to inaccuracies in the signal and noise subspaces obtained from the covariance matrix decomposition, which also requires extensive computation.
The maximum likelihood estimation method in \cite{60075} provided more stable DOA performance with low SNR and a small number of snapshots compared to subspace classification methods.

Techniques that integrate the time-domain and spatial-domain characteristics of signals had received considerable attention in recent years, particularly spatial spectrum estimation methods based on higher-order cumulants.
The literature \cite{zhang2015improved} constructed matrices utilizing fourth-order cumulants and performed eigenvalue decomposition, followed by MUSIC for DOA estimation, achieving more significant estimation accuracy compared to traditional second-order cumulants, although it introduced increased computational complexity.
To reduce this computational load, \cite{ahmed2014multiple} formed a higher-order matrix using fourth-order cumulants and then employed ESPRIT for angle estimation.
Furthermore, a high-resolution DOA estimation method that utilizes fourth-order accumulation to rapidly eliminate redundancy is proposed in \cite{10318118}. The redundant data in the fourth-order accumulation is reduced by a selection matrix in descending order, and then the redundancy in the vectorization process is eliminated by applying the vectorized form of the fourth-order cumulant matrix transformation.
Finally, DOA estimation is performed using the sparse representation method. In comparison with the conventional fourth-order cumulant method, this approach provides higher estimation accuracy and resolution, with enhanced capabilities for suppressing colored noise. Based on the above analysis, the DOA estimation method based on higher-order cumulants effectively suppresses both Gaussian and non-Gaussian noise and supports array expansion.

Most researchers initially focused on one-dimensional (1D) DOA estimation based on uniform linear array (ULA) structures. However, two-dimensional (2D) DOA estimation, which provides both azimuth and elevation angles of the source, has gained considerable attention due to its greater practical research significance. 
Joint singular value decomposition (SVD) of two cross-correlation matrices was proposed in \cite{4357952} to automatically pair and estimate 2D angles.
However, the SVD of higher-order covariance matrices involved high computational complexity. In \cite{1597167}, the azimuth and elevation angles were estimated by two separate 1D ESPRIT algorithms, with pair-matching achieved through the cross-correlation matrix of the two received signals from each ULA. Despite this, the use of two 1D parameter estimations and subsequent pairing resulted in high computational loads.
To resolve the issue of parameter paired, an enhanced propagator method with automatic parameter pairing based on two-parallel ULAs was introduced \cite{LI20123032}, showing improved estimation accuracy and reduced computational load compared to \cite{WU20031827}.
Due to its simpler array structure, larger array aperture and more accurate DOA estimation, 2D DOA estimation based on L-shaped arrays has become a significant research topic.
In \cite{1427919}, a method was proposed that transforms 2D DOA estimation of an L-shaped array into two independent 1D DOA estimations, thus eliminating the requirement for pair matching in parallel structures and reducing computational effort. Additionally, \cite{10292874} proposed a tensor-based iterative 2D DOA estimation method for L-shaped nested arrays to mitigate cross-terms caused by correlated co-array signals and noise components. This approach utilized high-order tensor decomposition to independently estimate azimuth and elevation angles, which were then effectively paired using the spatial cross-correlation matrix.

\subsection{Deep learning-based methods} 

With the advancement of deep learning (DL) \cite{li2023deep,xinyi2023machine,10505315}, DL-based DOA estimation methods have emerged as a new research hotspot. The DOA estimation problem can be transformed into a neural network classification issue, where the network learns the mapping relationship between the input data and the DOA to classify spatial angles. Alternatively, it can be approached as a neural network regression problem. Numerous studies have demonstrated that the computational efficiency of DL-based DOA estimation algorithms has significantly improved.

Fully connected neural network (FCNN) \cite{8400482, 9151162} has been widely used in DOA estimation and related fields, but its large number of matrix multiplication operations resulted in high computational costs. 
Consequently, convolutional neural network (CNN), which uses convolutional operations instead of matrix multiplication in FCNN, have gradually emerged in the field of DOA estimation. \cite{9779448} used CNN to estimate the noiseless covariance matrix of the generalized ULA and retrieved the DOA using root-MUSIC, thereby overcoming the grid mismatch issues encountered with grid-based methods.
Moreover, improved CNN-based networks have been introduced to further enhance the DOA estimation performance.
A multi-branch convolutional recurrent neural network with residual links and a weighted noise subspace network were introduced in \cite{10526225} to improve the covariance matrix estimation, subspace partitioning, and peak-finding processes.
Subsequently, multi-stage neural network-based DOA estimation was proposed to address the limitation of single neural networks in generalizing across all conditions. For example, a cascade neural network was proposed in \cite{chung2021off} with two stages: the first stage used a CNN to obtain an initial DOA value, and the second stage used an FCNN to generate a tuning vector representing the difference between the true DOA and the closest discretization angle. The final DOA estimation was then obtained by adjusting the initial DOA estimate using the tuning vector.
In \cite{10530534}, a cascade off-grid network consisting of an autoencoder composed of FC and a deep CNN with 2D convolutional layers was proposed, which used off-grid errors as labels to achieve off-grid DOA estimation based on its sparsity. The two-stage network discussed above facilitates in solving the mismatch caused by off-grid, thereby effectively improving DOA estimation performance.

Moreover, given that the array outputs are complex-valued in the DOA estimation systems,
complex valued neural network-based DOA estimation was suggested to extract the relevant DOA feature information from the complex-valued outputs of the arrays using complex parameters and complex arithmetic operations. 
Tan et al. \cite{10172255} proposed a complex-valued convolutional neural network with phase normalization to extract explicit phase information from intermediate complex-valued feature maps for estimating the unknown source DOAs.
A complex-valued residual angle estimation neural network containing complex linear and complex convolutional layers was introduced in \cite{10458139}, which combined the initial feature extraction module, the deep feature extraction module and the mapping module to perform DOA estimation.
However, the labeled datasets are usually difficult to collect in realistic DOA estimation systems. Therefore, some semi-supervised learning\cite{9449880}, unsupervised learning\cite{10379083}, and transfer learning \cite{9582746,10409285} methods have been proposed to resolve the problem of limited labeled datasets.
These methods can improve DOA estimation performance with less data and shorter training times compared to supervised learning-based methods.

\subsection{Motivations and contributions}

Although the methods analyzed above can better balance the relationship between resolution, estimation accuracy, and computational complexity to improve DOA estimation performance at low SNR and with a small number of snapshots. However, most of the current studies conduct experimental validation using ideal target incident angles while neglecting noise. 
The ideal angle is difficult to realize because noise always exists in real communication. Moreover, directly using the angles estimated from advanced DOA methods, like root-MUSIC or DL, to calculate weighted coefficients can result in certain performance loss. Therefore, we will focus on how to implement a more practical, low-cost, high-performance as well as high time-efficiency passive DOA estimator.
The critical contributions of this paper are summarized as follows:
\begin{enumerate}
    
    \item  On the basis of heterogeneous hybrid MIMO receive arrays, a multi-modal fusion DOA framework is established to achieve a low-latency, high-accuracy and more realistic DOA measurements. The framework involves three main steps: 1) FD subarray and all subarray groups of $\rm{H}^2$AD array form a rough DOA value and candidate angle sets, respectively, via the Root-MUSIC algorithm. 2) The true angle classes are distinguished from the candidate angle set using the clustering method based on the initial sample points provided by the FD subarray.
    3) The two-part true angle sets are fused to achieve an improved DOA performance.

    \item Based on the framework described above, an effective clustering methods, named global maximum similarity (GMaxCS), is proposed to attain more accurate true solutions from the candidate angle sets. These methods have the advantage of directly reflecting the angles between signal direction vectors and significantly enhancing the differentiation of direction information. A co-learning-assisted multi-modal iteration weighted fusion (MM-IWF)-based method is then proposed to fuse the two-part true angles using the estimated values.
    The clustering and fusion process described above results in two DOA estimation methods for eliminating phase ambiguity, named MM-IWF-GMinD and MM-IWF-GMaxCS. The weighted fusion coefficients can be more accurately obtained through iterative operations, making the final DOA estimation closer to the ideal values. Additionally, by utilizing the estimated angles to design the fusion coefficients, a more practical passive DOA estimator is achieved. Experimental results show that the proposed MM-IWF-GMaxCS outperforms MM-IWF-GMinD.

    \item To further achieve a higher-accuracy DOA estimation, a co-learning-assisted multi-modal fusion network (fusionNet)-based on a three-layer FCNN is introduced to aggregate the inferred true angles of all subarray groups and a coarse DOA angle of FD subarray. The fusion weights, adaptively learned by fusionNet, are not directly correlated with the ideal DOA angles. Following the same pattern as the previous methods, the two novel corresponding approaches are called as follows: MM-fusionNet-GMinD and MM-fusionNet-GMaxCS. To validate the effectiveness of these approaches, the hybrid Cramér-Rao Lower Bound (CRLB) is utilized as a baseline. Simulation results illustrate that our proposed four clustering-fusion approaches–based co-learning DOA estimation can approach the desired DOA performance. Also, the proposed MM-fusionNet-GMinD and MM-fusionNet-GMaxCS show superior DOA estimation performance compared to the proposed MM-IWF-GMinD and MM-IWF-GMaxCS, especially in ultra low SNR range.

\end{enumerate}

The remaining parts of the study are arranged as follows. Section \ref{sec_sys} shows the system model. A co-learning-assisted fusion DOA framework for heterogeneous hybrid structure is introduced in Section \ref{sec_proposed}. In Section \ref{sec_CNN}, 
the DOA clustering and iteration weighted fusion methods for the proposed frameworks are detailed. Section \ref{FUSION} describes the high-performance fusionNet developed for DOA estimation. The CRLB of system model and computational complexity of our proposed approach are analyzed in Section \ref{sec_perf}. The simulation results are provided in Section \ref{sec_simu} and Section \ref{sec_con} offers conclusions. 

\emph{Notations:} In this paper, uppercase letters and lowercase letters in bold typeface (i.e., $\mathbf{B}$, $\mathbf{b}$) are matrices and vectors, respectively. Signs $(\cdot)^H$, $(\cdot)^T$,$\|\cdot\|$, $|\cdot|$, and $\Re\{\cdot\}$ denote conjugate transpose, transpose, norm, modulus, and real part operations, respectively.

\section{System Model}\label{sec_sys}

\begin{figure}[!htb]
	\centering
	\includegraphics[width=4.8in]{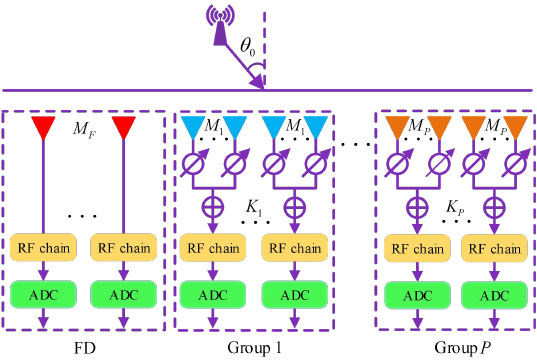}
	\caption{A heterogeneous hybrid massive MIMO receiver.}
 \label{fig: sub_figure0}
\end{figure}

Figure~\ref{fig: sub_figure0} shows a heterogeneous hybrid massive MIMO receiver with a FD subarray and a $\rm{H}^2$AD array, where $\rm{H}^2$AD consists of $P$ groups, group $p$ contains $K_p$ subarrays, and each subarray has $M_p$ antennas. 
\begin{align} \label{11}
	{{M}_{{\rm{H}}^{2}AD}}=\sum\limits_{p=1}^{P}{{{N}_{p}}}=\sum\limits_{p=1}^{P}{{{K}_{p}}}{{M}_{p}},
\end{align}
where $M_1\neq M_2 \neq\cdots\neq M_P$ and the values of $M_1, M_2, \cdots, M_P$ are prime numbers.

Assume there is an emitted signal incident from $\theta_0$, 
the received signal vector of FD subarray and $p$-th group are respectively expressed as
\begin{align} 
	\mathbf{y}_{F}(n)=\mathbf{a}_{F}(\theta_0)e(n)+\mathbf{v_F}(n),
\end{align}

\begin{align}	\mathbf{y}_p(n)=\mathbf{\Xi}_{A,p}^H\mathbf{a}_p(\theta_0)e(n)+\mathbf{v_p}(n),
\end{align}
where $e(n)$ is the baseband signal, $n=1,2,\cdots,H$, $H$ is the number of snapshots. $\mathbf{v_F}(n)\sim\mathcal{C}\mathcal{N}(0,\sigma^2_v\mathbf{I})$ and $\mathbf{v_p}(n)=\left[v_1(n), v_2(n), \ldots, v_{K_p}(n)\right]^T\in\mathbb{C}^{K_p \times 1}$ are the additive white Gaussian noise (AWGN) vector, $\mathbf{a}_{F}\in\mathbb{C}^{M\times 1}$ and  $\mathbf{a}_p(\theta_0)\in\mathbb{C}^{N_p \times 1}$ are the array manifold vector defined as
 
\begin{align}
	\mathbf{a}_{F}\left(\theta_0\right)=\left[e^{j {2 \pi\varphi(1)}}, \cdots,e^{j {2 \pi}{\varphi(m_1)}},\cdots, e^{j {2 \pi}{\varphi(M)}}\right]^T,
\end{align}

\begin{align}
	\mathbf{a}_{p}\left(\theta_0\right)=\left[e^{j {2 \pi\varphi(1)}}, \cdots,e^{j {2 \pi}{\varphi(m_2)}},\cdots, e^{j {2 \pi}{\varphi(N_p)}}\right]^T,
\end{align}
where 
\begin{align}
	\varphi(m_1)= \frac{({m_1}-1) d \sin \theta_0}{\lambda} ,m_1=1,2,\cdots,M
\end{align}
\begin{align}
	\varphi(m_2)= \frac{({m_2}-1) d \sin \theta_0}{\lambda} ,m_2=1,2,\cdots,N_p
\end{align}
where $\lambda$ is carrier frequency wavelength. The phase reference point is established at the leftmost side of the array and $d=\frac{\lambda}2$.

And a block diagonal matrix $\mathbf{\Xi}_{A,p}$ is expressed by 
\begin{align}
{\mathbf{\Xi _{A,p}}} = \left[ {\begin{array}{*{20}{c}}
{{\varepsilon _{A,p,1}}}&0& \cdots &0\\
0&{{\varepsilon _{A,p,2}}}& \cdots &0\\
 \vdots & \vdots & \ddots & \vdots \\
0&0& \cdots &{{\varepsilon _{A,p,K}}}
\end{array}} \right],
\end{align}
where $\mathbf{\varepsilon}_{A,p,k}$ is the $k$-th element denoted as 
\begin{align}
	\mathbf{\varepsilon}_{A,p,k}=\frac{1}{\sqrt{M_p}}\left[e^{j\Upsilon_{p,k,1}},e^{j\Upsilon_{p,k,2}},\cdots,e^{j\Upsilon_{p,k,M_p}}\right]^T,
\end{align}
where $\Upsilon_{p,k,m}$ is the analog beamforming phase.

\section{Proposed Co-learning-Assisted Fusion DOA Framework For Heterogeneous Hybrid Structure}\label{sec_proposed}

In this section, a co-learning-assisted fusion DOA framework for a heterogeneous hybrid massive MIMO receiver is introduced to expedite the clustering of true and false solution classes and enhances the accuracy of weighted coefficients through iterative processing and FusionNet, thereby improving DOA estimation capabilities. 

\begin{figure}[!htb]
	\centering
	\includegraphics[width=4.8in]{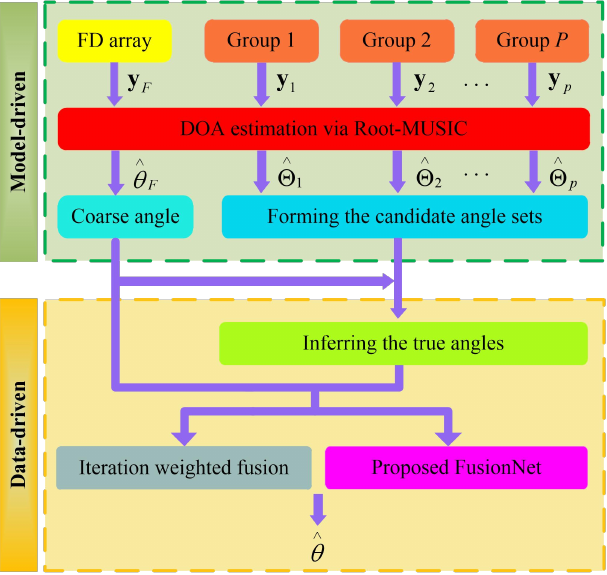}
	\caption{Co-learning-assisted fusion DOA framework.}
 \label{fig: frame}
\end{figure}

As shown in Figure~\ref{fig: frame}, the proposed co-learning-assisted fusion DOA framework contains three major stages: 1) The coarse DOA value of FD subarray and the candidate angle sets of all subarray groups are obtained via Root-MUSIC algorithm; 
2) distinguish the true angle classes from the set of candidate angles by two introduced clustering method based on the initial sample points offered by the FD subarray; 
3) fuse the two-part true angles to obtain the final DOA value through the proposed iteration weighted fusion (IWF)-based method and fusionNet.

Then the spatial spectrum corresponding to FD subarray is given by 
\begin{align} \label{wewew}
	S_{F}(\theta)=\frac{1}{\left\|\mathbf{a}_{F}^H(\theta)\mathbf{U}_N\right\|^2},
\end{align}

The denominator portion of $S_{F}(\theta)$ can be defined by 
\begin{equation}
\begin{aligned} 	
s_F(\theta)&=\mathbf{a}_{F}^H(\theta)\mathbf{U}_N\mathbf{U}_{N}^H\mathbf{a}_{F}(\theta) \\ 
& =\mathbf{a}_{F}^H(\frac{1}{{{z_F}}})\mathbf{U}_N\mathbf{U}_{N}^H\mathbf{a}_{F}(z_{F})\\
&\triangleq s_F(z_{F})=0,
\end{aligned}
\end{equation}
The polynomial equation $s(z_{F})$ has $2(M_{F}-1)$ roots, thus the root closest to the unit circle represents the intended DOA direction.

\begin{align} \label{FDeeee}
	\hat{\theta}_{F}=\arcsin \left(\frac{\lambda}{2 \pi d} \arg z_{F}\right)
\end{align}

For $\rm{H}^2$AD array, spatial spectrum of the virtual antenna array is calculated by

\begin{align} \label{}
	S_{\rm{H}^2AD}(\theta)=\frac{1}{\left\|r_p(\theta)\right\|^2\left\|\mathbf{a}_{M_p}^H(\theta)\mathbf{E}_N\right\|^2},
\end{align}
where 

\begin{align} \label{}
	r_p\left(\theta\right)
	=\frac{1-e^{j \frac{2 \pi}{\lambda} M_p d \sin \theta}}{1-e^{j \frac{2 \pi}{\lambda} d \sin \theta}}
\end{align}

Let define $\mathbf{B}=\mathbf{E}_N\mathbf{E}_N^H$, then

\begin{equation}\label{ppppp}
\begin{aligned} 	
{{S}_{{{\text{H}}^{\text{2}}}\text{AD}}}^{-1}(\theta )
=\frac{2-{{e}^{-j\frac{2\pi }{\lambda }{{M}_{p}}d\sin \theta }}-{{e}^{j\frac{2\pi }{\lambda }{{M}_{p}}d\sin \theta }}}{2-{{e}^{-j\frac{2\pi }{\lambda }d\sin \theta }}-{{e}^{j\frac{2\pi }{\lambda }d\sin \theta }}}\cdot  \\ 
\sum\limits_{i_1=1}^{{{K}_{p}}}{\sum\limits_{i_2=1}^{{{K}_{p}}}{{{e}^{-j\frac{2\pi }{\lambda }(i_1-1){{M}_{p}}d\sin \theta }}}}{{\mathbf{B}}_{i_1 i_2}}{{e}^{j\frac{2\pi }{\lambda }(i_2-1){{M}_{p}}d\sin \theta }}
\end{aligned}
\end{equation}

Via Root-MUSIC \cite{1993The}, the angle obtained from the peak of $S_{\rm{H}^2AD}(\theta)$ is the desired DOA value. And the polynomial equation in $S_{\rm{H}^2AD}(\theta)$ is defined by
\begin{equation}\label{p}
\begin{aligned} 	s(\theta)&=r_p^H(\theta)\mathbf{a}_{M_p}^H(\theta)\mathbf{E}_N\mathbf{E}_N^H\mathbf{a}_{M_p}(\theta)r_p(\theta)\triangleq s(z)\\
	&=\frac{2-z^{-1}-z}{2-z^{-\frac{1}{M_p}}-z^{\frac{1}{M_p}}}\sum_{i_1=1}^{K_p}\sum_{i_2=1}^{K_p}z^{-(i_1-1)}\mathbf{B}_{i_1 i_2}z^{i_2-1}\\
   &\triangleq s(\eta)=0,
\end{aligned}
\end{equation}

And 

\begin{align} \label{z}
	z=e^{j\eta_p},
\end{align}

\begin{align} \label{}
	\eta_p=\frac{2 \pi}{\lambda} M_p d \sin \theta,
\end{align}

The polynomial equation $s(\theta)$ has $2K_p-2$ roots, i.e., ${Z}_{R}=\left\{{z}_l, l \in\left[1, 2 K_p-2 \right]\right\}$. Moreover, DOA estimation sets can further be expressed as 

\begin{align} \label{}
	\hat{\Theta}_{R }=\left\{\hat{\theta}_l, l \in \left[1, 2 K_p-2 \right] \right\},
\end{align}
where
\begin{align} \label{}
	\hat{\theta}_l=\arcsin \left(\frac{\lambda \arg z_l}{2 \pi M_p d}\right)
\end{align}
Thus, the estimation value $\hat\theta_{p}$ of $p$-th group is 
\begin{align} \label{}
	\hat{\eta}_p=\frac{2\pi}{\lambda} M_pdsin\hat{\theta_{p}}
\end{align}

Considering that the function $f(\hat{\eta}_p)=f(\hat{\eta}_p+2\pi q)$ with $2\pi$ period, thus the feasible sets containing $M_p$ solutions of $p$-th group can be extended as follows

\begin{align} \label{THETA_q}
	\hat{\Theta}_{p}=\left\{\hat{\theta}_{p, q_p}, q_p \in\{1,2, \cdots, M_p\}\right\}
\end{align}
where
\begin{align} \label{}
	\hat{\theta}_{p, q_p}=\arcsin \left(\frac{\lambda\left(\arg (e^{j\hat{\eta}_p})+2 \pi q\right)}{2 \pi M_p d}\right) .
\end{align}
Integrating all $P$ groups has
\begin{align}\label{angleCandAll}
	\begin{split}
		\left \{
		\begin{array}{ll}
			\frac{2\pi}{\lambda}M_1dsin\theta_{1,q_1}=\hat{\eta}_1+2\pi q_1\\
			\frac{2\pi}{\lambda}M_2dsin\theta_{2,q_2}=\hat{\eta}_2+2\pi q_2\\
			\quad\quad\quad\vdots\quad\quad\quad\quad\quad\quad\quad\vdots\\
			\frac{2\pi}{\lambda}M_Pdsin\theta_{P,q_P}=\hat{\eta}_P+2\pi q_P
		\end{array}
		\right.
	\end{split}
\end{align}
where $q_p\in\{1,2,\cdots,M_p\}$ is the ambiguity coefficient. Thus, the candidate set of $\rm{H}^2$AD can be represented as
\begin{align} \label{}
	\hat{\Theta}=\left\{\hat{\Theta}_{1}, \hat{\Theta}_{2} \cdots, \hat{\Theta}_{P}\right\}
\end{align}

Based on the above analysis, the candidate set $\hat{\Theta}$ contains $\sum\limits_{p = 1}^P {{M_{_p}}}$ solution and each candidate set ${\hat{\Theta}_p}$ has a true solution and $M_{_p}-1$ false angles, thus it is a vital issue to eliminate the pseudo-solutions. When noise is negligible, the true angle class and false angles class can be respectively expressed as 


\begin{align} \label{bb}	
{\hat{\theta}_{c,1}} \approx {\hat{\theta}_{c,2}} \approx  \cdots  \approx {\hat{\theta}_{c,P}} \approx {\theta _0}
\end{align}

\begin{align} \label{DD}
{\hat{\theta}_{f,1,m}} \ne {\hat{\theta}_{f,2,m}} \ne  \cdots  \ne {\hat{\theta}_{f,P,m}} \ne {\theta _0}
\end{align}
where 
\begin{align} \label{222DD}
\hat{\theta}_{f,p,m}=\theta_0+\chi_{p,m},m\in \left[1,M_p-1\right]
\end{align}
Each group has a different number of antennas, resulting in a difference about $\chi_{p,m}$ value, therefore, equation (\ref{DD}) is reasonable.

Then, the true-solution classes is represented as 
\begin{equation}\label{eeeee}
	\hat{\Theta}_{c}=\left\{ \hat{\theta}_{c,1},\hat{\theta}_{c,2},\cdots,\hat{\theta}_{c,P} \right\}
\end{equation}
where $\hat{\theta}_{c,p}$ denotes the predicted true solution of the $p$-th subarray group.

\section{Proposed DOA Clustering and Iteration Weighted Fusion Methods}\label{sec_CNN}
\begin{figure}[!htb]
	\centering 
	\includegraphics[width=5.0in]{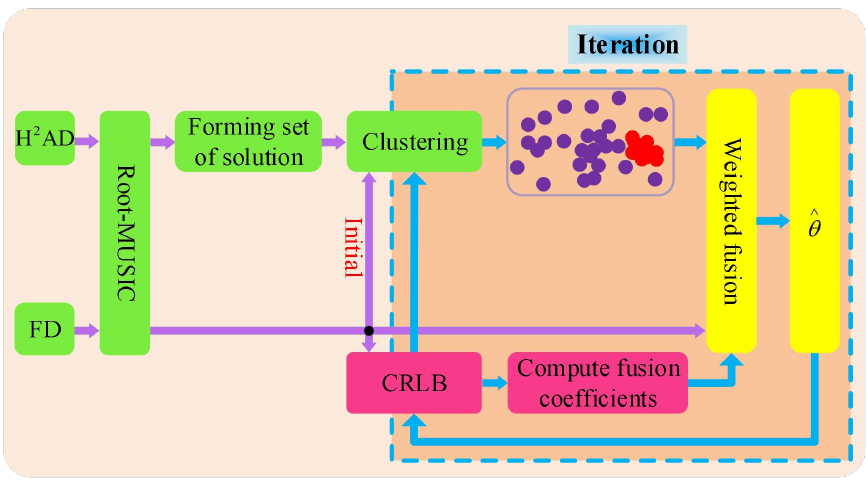}
	\caption{Co-learning-assisted iteration weighted fusion for DOA estimation. }
 \label{fig: sub_figure1}
\end{figure}

In this section, two high-performance clustering methods, 
GMaxCS and GMinD, is introduced to accelerate the clustering of true and false solution classes. Based on this, a fusion method, i.e., Co-IWF, is proposed to obtain more precise DOA estimation, shown in Figure \ref{fig: sub_figure1}. The above process forms two DOA estimation methods: Co-ITW-GMinD and Co-ITW-GMaxCS.

\subsection{Proposed GMaxCS Clustering.}
Given that cosine similarity directly reflects the angle between signal direction vectors, it significantly enhances the differentiation of direction information, thereby improving the accuracy of DOA estimation. Based on this, we proposed a high-accuracy clustering method, called GMaxCS.

When the initial DOA value $\hat{\theta}_{F}$ and the candidate set $\hat{\Theta}_{p}$ of $p$-th group is acquired. We can define the vector

\begin{align}
  & u  =\left( cos\overset{\wedge }{\mathop{\theta }}_{F}\,  ,sin\overset{\wedge }{\mathop{\theta }}_{F}\,  \right) \ \\ 
 & v=\left( cos{{\overset{\wedge }{\mathop{\theta }}\,}_{p,{{q}_{p}}}},sin{{\overset{\wedge }{\mathop{\theta }}\,}_{p,{{q}_{p}}}} \right)  
\end{align}

Then, the cosine\_similarity can be defined by

\begin{equation}
    \cos\_\rm{sim} \left\langle u,v \right\rangle =\frac{u\cdot v}{\left\| u \right\|\left\| v \right\|}
\end{equation}

Thus, the true angle ${\overset{\wedge }{\mathop{\theta }}\,}_{c,p}$ is obtained by searching the direction with maximum similarity 

\begin{equation}\label{rtryhtj11}
   {{\overset{\wedge }{\mathop{\theta }}\,}_{c,p}}=\underset{{{\overset{\wedge }{\mathop{\theta }}\,}_{p,{{q}_{p}}}}\in {{\overset{\wedge }{\mathop{\Theta }}\,}_{P}}}{\mathop{\arg \max }}\,\cos\_\rm{sim} \left\langle u,v \right\rangle 
\end{equation}

\subsection{Proposed Co-IWF Method.}

\begin{figure}[!htb]
	\centering
	\includegraphics[width=4.3in]{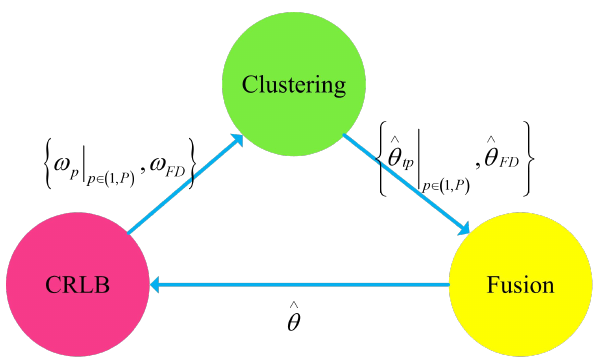}
	\caption{Proposed iteration weighted fusion method.}
 \label{fig: sub_figure3}
\end{figure}

Figure \ref{fig: sub_figure3} shows iterative process for the proposed IWF method, and the detailed procedure is described as follows. The FD subarray gives an initial sample point for each group to
infer the true solution class from the $p$-th candidate set $\hat{\Theta}_{p}$. Also, it is used to calculate the initial fusion coefficients. The global minimum distance clustering (GMinD) \cite{shu2024newheterogeneoushybridmassive} is expressed by

\begin{align} \label{gsProbnnn}	
\hat{\theta}_{c,p}=\underset{\hat{\theta}_{p,q_p}\in\hat{\Theta}_p}{\arg \min }\left\|\hat{\theta}_{F}-\hat{\theta}_{p,q_p}\right\|^2
\end{align}

Then the desired DOA value can be obtain by iteration fusing the two-part true solution 
\begin{align} \label{thetaEstConbine111}
	\hat{\theta}^{(i+1)}= w_{F}(\hat{\theta}^{(i)})\hat{\theta}_{F} + \sum_{p=1}^{P} w_{p}(\hat{\theta}^{(i)})\hat{\theta}_{c,p} 
\end{align}
where $\hat{\theta}^{(0)}=\hat{\theta}_{F}$, $i$ denotes the number of iterations.

According to the mean square error of the $\hat{\theta}^{(i)}$, the problem is transformed into Eq.(\ref{eqn222})

\begin{equation} \label{eqn222}
  \begin{aligned}
   & \underset{\begin{smallmatrix} 
 {{w}_{p}}(\hat{\theta }^{(i)}) \\ 
 {{w}_{F}}(\hat{\theta }^{(i)}) 
\end{smallmatrix}}{\mathop{\min }}\,\sum\limits_{p=1}^{P}{{{w}_{p}}}^{2}{{(\hat{\theta }^{(i)})}}CRL{{B}_{p}}(\hat{\theta }^{(i)})+w^{2}_{F}(\hat{\theta}^{(i)})CRL{{B}_{F}}(\hat{\theta }^{(i)})  \\
   &  \quad s.t.~~~~~{{w}_{F}}(\hat{\theta }^{(i)})+\sum\limits_{p=1}^{P}{{{w}_{p}}}(\hat{\theta }^{(i)})=1,  
\end{aligned}
\end{equation}

\noindent where weighted fusion coefficients $w_{F}(\hat{\theta}^{(i)})$ and $w_{p}(\hat{\theta}^{(i)})$ is respectively expressed by
\begin{align} \label{wqFinal1}
{{w}_{F}}(\hat{\theta }^{(i)})=\frac{CRLB_{F}^{-1}(\hat{\theta }^{(i)})}{CRLB_{F}^{-1}(\hat{\theta }^{(i)})+\sum\limits_{p=1}^{P}{C}RLB_{p}^{-1}(\hat{\theta }^{(i)})}
\end{align}

\begin{align} \label{wqFinal2}
 w_{p}(\hat{\theta}^{(i)})=\frac{{C R L B_{p}^{-1}}(\hat{\theta}^{(i)})}{{C R L B_{F}^{-1}(\hat{\theta}^{(i)})+\sum\limits_{p=1}^{P}{C}RLB_{p}^{-1}(\hat{\theta }^{(i)})}}
\end{align}
where $C R L B_{F}(\hat{\theta}^{(i)})$ and $C R L B_{p}(\hat{\theta}^{(i)})$ are expressed by Eq.(\ref{CRLB11}) and Eq.(\ref{CRLB2}), respectively.

\begin{align}\label{CRLB11}
C R L B_{F}(\hat{\theta}^{(i)}) = \frac{{{\lambda ^2}}}{{8H{\pi ^2}\mathbf{SNR}{{\cos }^2}{(\hat{\theta}^{(i)})}\mathop {{d^2}}\limits^ - }}
\end{align} 

\begin{equation}\label{CRLB2}
		CRLB_{p}(\hat{\theta}^{(i)}) = \frac{\frac{1}{8H{\pi ^2}\mathbf{SNR}{{\cos }^2}{(\hat{\theta}^{(i)})}} \lambda ^2{M_p}\Psi^{(i)}}{\left[{\frac{{{\left\| {{r_p}({\theta _0})} \right\|}^4}M_p^2 K_p^2\left( {K_p^2 - 1} \right){d^2}}{{12\Psi^{(i)}}} + \frac{{M_p{K_p}}}{\Psi^{(i)}}\left( {{{\left\| {{r_p}({\hat{\theta}^{(i)}})\varsigma^{(i)}} \right\|}^2} + {K_p}\Re\left\{ {{r_p}^2({\hat{\theta}^{(i)}})\varsigma^{(i)} } \right\}} \right) }
 \right]}
\end{equation}

\noindent where
\begin{align}
	\varsigma^{(i)} = \sum_{m=1}^{M_p}\left(m-1\right)d e^{-j\frac{2\pi}{\lambda}(m-1)d\sin{(\hat{\theta}^{(i)})}}
\end{align}
\begin{align}
	\Psi^{(i)}  = {M_p} + {K_p}\mathbf{SNR}{\left\| {{r_p}{(\hat{\theta}^{(i)})}} \right\|^2}
\end{align}

The overall algorithm of the above method, Co-IWF-GMinD and Co-IWF-GMaxCS, is shown in Algorithm \ref{alg:GS1}.


\begin{algorithm}[t]
	\caption{Proposed Co-IWF-GMinD and Co-IWF-GMaxCS.}\label{alg:GS1}
	\begin{algorithmic}
		\STATE 
  \STATE {\textbf{Input:}} $\mathbf{y}_{F}(n)$ and $\mathbf{y}_p(n)$, $p=1,2,\cdots,P$
		\STATE \hspace{0.5cm} Calculate the $\hat{\theta}_{F}$, weighted fusion coefficients $w_{F}(\hat{\theta}^{(0)})$ and $w_{p}(\hat{\theta}^{(0)})$.  
  \STATE \hspace{0.5cm} Calculate $\hat{\theta}_{c,p}^{(i)}$ by (\ref{rtryhtj11}) or (\ref{gsProbnnn}).
  \STATE \hspace{0.5cm} Set $i=0$, relative error $\sigma$.
		\STATE  \textbf{repeat} 
   \STATE \hspace{0.5cm} $i=i+1$.
   
    \STATE \hspace{0.5cm} Update $w_{F}(\hat{\theta}^{(i)})$ and $w_{p}(\hat{\theta}^{(i)})$ by (\ref{wqFinal1}) and (\ref{wqFinal2}).
   \STATE \hspace{0.5cm} Update $\hat{\theta}^{(i+1)}$ by (\ref{thetaEstConbine111}).
  \STATE \textbf{Until}$\left\| {\hat{\theta}^{(i+1)}-\hat{\theta}^{(i)}} \right\|\le \sigma $.
		\STATE {\textbf{Output:}} $\hat{\theta}^{(i+1)}$
	\end{algorithmic}
\end{algorithm}

\section{Proposed High-performance Fusion Network For DOA Estimation}\label{FUSION}
\begin{figure}[!htb]
	\centering
	\includegraphics[width=4.8in]{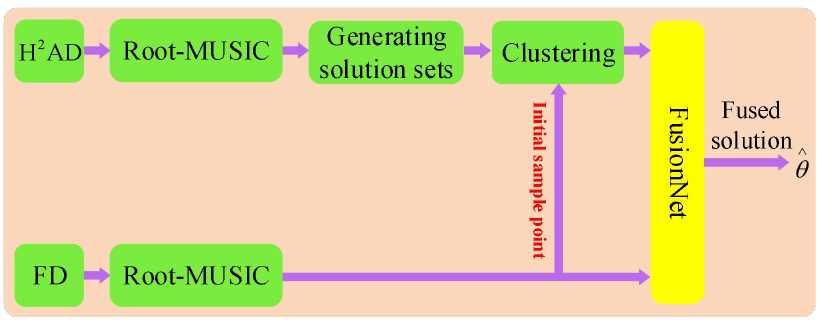}
	\caption{Proposed fusionNet–based co-learning DOA estimation.}
 \label{fig: fusionnet5}
\end{figure}

In this section, a fusionNet–based co-learning (Co-fusionNet) DOA estimation is proposed, shown in Figure \ref{fig: fusionnet5}. The fusionNet utilizes a three-layer fully connected neural network (FCNN), consisting of two hidden layer and an output layer, for the fusion of the two-part true solutions, 
thereby providing a higher DOA estimation accuracy. 
Moreover, the fusion weights learned by the network are not correlated with the ideal DOA angles, making proposed fusionNet better suited for DOA estimation in real-world scenarios.

\subsection{FusionNet Structure}
\begin{figure}[!htb]
	\centering
	\includegraphics[width=4.2in]{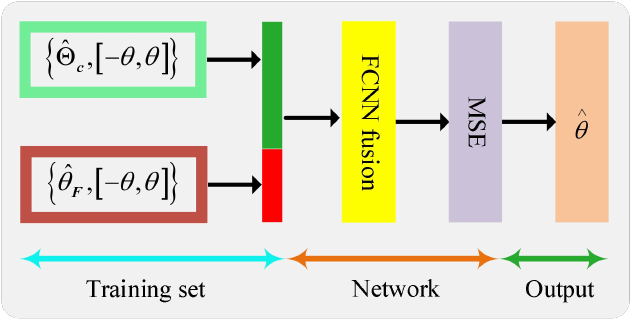}
	\caption{The fusionNet implementation process.}
 \label{fig: process}
\end{figure}

The inferred true angles can be aggregated by the introduced fusionNet to obtain the final DOA estimation. The detailed implementation process is shown in Figure \ref{fig: process}. The proposed adaptive fusionNet processes the true solutions of the concatenated $\rm{H}^2$AD and FD subarrays through a two-layer FCNN fusion. The following equation is utilized to acquire the expected DOA value:

\begin{equation}\label{WWWWQ}
\mathop \theta \limits^ \wedge   = FCNN\left( {C\left( {{{\mathop \Theta \limits^ \wedge  }_c},{{\mathop \theta \limits^ \wedge  }_F}} \right); \omega } \right)   
\end{equation}
where $C\left( {{{\mathop \Theta \limits^ \wedge  }_c},{{\mathop \theta \limits^ \wedge  }_F}} \right)$ is a new vector spliced by the true solution of $\rm{H}^2$AD and rough estimate value of FD subarray. $\omega$ is the learned parameters of the fusionNet. 

According to the characterization of the FCNN layer, Eq.(\ref{WWWWQ}) is transformed to
\begin{equation}\label{WWWWiiQ}
\begin{aligned} 	
\mathop \theta \limits^ \wedge=\mathbf{W} \cdot C\left( {{{\mathop \Theta \limits^ \wedge  }_c},{{\mathop \theta \limits^ \wedge  }_F}} \right) + \mathbf{b},
\end{aligned}
\end{equation}
where $\mathbf{W}$ is the weight matrix learned through the training process of the FCNN, $\mathbf{b}$ is the bias vectors. During the training process, the fusionNet adaptively adjusts weights through a back-propagation algorithm to optimize the performance of the fusion metric results.


\subsection{Date Management and Loss Function}
Based on the above analysis, the true solution set $\hat{\Theta}_{c}=\left\{ \hat{\theta}_{c,1},\hat{\theta}_{c,2},\cdots,\hat{\theta}_{c,P} \right\}$ of the $\rm{H}^2$AD array generated by co-learning-assisted clustering as well as a coarse DOA estimation $\hat{\theta}_{F}$ of the FD sub-array can be used as the input to the network. Assuming that the angle of the training set is chosen to be within $ {\left[ { - \theta ,\theta } \right]}$, where $\theta\in[0^\circ,90^\circ]$. 
The training dataset and its labels are generated based on different angles corresponding to different DOA estimations.
Thus, the entire training set of the proposed fusionNet is denoted by $ {\left\{ {\left( {{{\hat \Theta }_c},{{\hat \theta }_F}} \right)\left| {_{_{\left[ { - \theta ,\theta } \right]}}} \right.,\left[ { - \theta ,\theta } \right]} \right\}}  $.

Considering that mean squared error (MSE) provides a smooth and easy-to-compute measure of prediction error, achieving higher prediction accuracy in regression tasks. Therefore, the MSE is utilized as a loss function to optimize the model during the training process in our research. The MSE loss function is defined as follows:

\begin{equation}
    \mathcal{L_{MSE}} = \frac{1}{Q}{\sum\limits_{q = 1}^Q {\left( {{{\mathop \theta \limits^ \sim  }_q} - \theta } \right)} ^2}
\end{equation}
where $Q$ represents the number of training datasets, ${{{\mathop \theta \limits^ \sim  }_q}}$ and $\theta$ are the predicted DOA value and the true label, respectively.
We train the model with the aim of finding the parameters that minimize the loss function. By minimizing the $\mathcal{L_{MSE}}$, the predicted DOA value ${{{\mathop \theta \limits^ \sim  }_q}}$ will be as close as possible to the true value $\theta$, thus improving the model's prediction accuracy and performance.

The entire algorithm of our proposed fusionNet–based co-learning DOA estimation,named Co-fusionNet-GMinD and Co-fusionNet-GMaxCS, is described in Algorithm \ref{alg:GS11}.

\begin{algorithm}[t]
	\caption{Proposed Co-fusionNet-GMaxCS and Co-fusionNet-GMinD.}\label{alg:GS11}
	\begin{algorithmic}
		\STATE 
		\STATE {\textbf{Input:}} $\mathbf{y}_{F}(n)$ and $\mathbf{y}_p(n)$, $p=1,2,\cdots,P$
            \STATE \hspace{0.5cm} Use the root-MUSIC method for $\mathbf{y}_{F}(n)$ to calculate the $\hat{\theta}_{F}$.
		\STATE \hspace{0.5cm} \textbf{for} $p=1,2,\cdots,P$ \textbf{do},
		\STATE \hspace{1cm} Calculate the candidate solution set $\hat{\Theta}_{p}$ for $\mathbf{y}_p(n)$  via the root-MUSIC method.
		\STATE \hspace{0.5cm} \textbf{end for}
            \STATE \hspace{0.5cm} Solve the problem (\ref{rtryhtj11}) or (\ref{gsProbnnn}) to obtain true angle set $\hat{\Theta}_{c}$.
            \STATE \hspace{0.5cm} Fuse the coarse angle $\mathbf{y}_{F}(n)$ and true-solution sets $\left\{ \hat{\theta}_{c,1},\hat{\theta}_{c,2},\cdots,\hat{\theta}_{c,P} \right\}$ via the proposed FusionNet to obtain the final $\hat{\theta}$.
		\STATE {\textbf{Output:}} $\hat{\theta}$
	\end{algorithmic}
\end{algorithm}

\section{Theoretical Analysis}\label{sec_perf}
This section provides the theoretical characterization analysis for a heterogeneous hybrid massive MIMO
structure and the complexity of the proposed approaches.

\subsection{CRLB}
A lower bound of the variance on any unbiased DOA estimatior is provided via the CRLB. Hence, the corresponding CRLB for heterogeneous hybrid massive MIMO receiver is described in this subsection as a baseline to assess the performance of the proposed DOA estimation-based methods.

The hybrid CRLB is expressed by
\begin{align} \label{FFFFSSSS}
	\sigma_{\theta_0}^2 \geq \frac{1}{H}\mathbf{F}^{-1}
\end{align}
where $\mathbf{F}$ is the Fisher information matrix of heterogeneous hybrid massive MIMO receiver, expreesed by 
\begin{align} \label{}
	\mathbf{F} =\mathbf{F}_{F}+\sum_{p=1}^{P}\mathbf{F}_{p}
\end{align}
where $\mathbf{F}_F$ and $\mathbf{F}_{p}$ are the Fisher information matrix of FD subarray and the $p$-th group, respectively, expreesed by Eq.(\ref{FIMFD}) and Eq.(\ref{FIMHAD1})
\begin{align}\label{FIMFD}
\mathbf{F}_{F} = \frac{{8{\pi ^2}\mathbf{SNR}{{\cos }^2}{\theta _0}\mathop {{d^2}}\limits^ - }}{{\lambda ^2}}
\end{align}

\begin{equation}\label{FIMHAD1}	
 \mathbf{F}_{p}= \frac{{8{\pi ^2}\mathbf{SNR}^2}{{\cos }^2}{\theta _0}}{{{\lambda ^2}{M_p}\Psi }}\left[ {\frac{1}{{12}}{{\left\| {{r_p}({\theta _0})} \right\|}^4}M_p^2K_p^2\left( {K_p^2 - 1} \right){d^2} + \frac{{M_p{K_p}}}{\Psi}\left( {{{\left\| {{r_p}({\hat{\theta}})\varsigma} \right\|}^2} + {K_p}\Re\left\{ {{r_p}^2({\hat{\theta}})\varsigma } \right\}} \right) } \right],
\end{equation}

Thus, the closed-form expression of the CRLB is given by
by (\ref{FIMHAD111}).

\begin{equation}\label{FIMHAD111}
    CRLB = \frac{1}{H}{{\bf{F}}^{ - 1}} = \frac{{\frac{{{\lambda ^2}}}{{8H{\pi ^2}{\bf{SNR}}{{\cos }^2}{\theta _0}}}}}{{\left[ {\mathop {{d^2}}\limits^ -   + \sum\limits_{p = 1}^P {\left( {\frac{{{{\left\| {{r_p}({\theta _0})} \right\|}^4}{M_p}K_p^2\left( {K_p^2 - 1} \right){d^2}}}{{12\Psi }} + \frac{{{K_p}}}{{{\Psi ^2}}}\left( {{{\left\| {{r_p}({\theta _0})\varsigma } \right\|}^2} + {K_p}\Re \left\{ {{r_p}^2({\theta _0})\varsigma } \right\}} \right)} \right)} } \right]}}
\end{equation}

\subsection{Computational Complexity}\label{Complexity}
As illustrated in Figure~\ref{fig: frame}, the proposed framework has three main stages and the computational complexity of the first and second parts is the identical. The computational complexity of the first part is from calculating the Root-MUSIC algorithm for FD subarrays and all subarray groups, so it is denoted as
$\mathcal{O}\left({{M}^{3}}+8{{M}^{2}}+ML\left( 2M+3 \right)-8N+1 \right)$ and $\mathcal{O}\left( \sum\limits_{p=1}^{P}{{{K}_{p}}}{{N}_{p}}+4\left( {{K}_{p}}-1 \right)\left( 2{{K}_{p}}+{{N}_{p}}{{K}_{p}}-2 \right) \right)$, respectively. The second part of the proposed GMaxCS and GMinD involve different specific operations, but they have the same complexity of $\mathcal{O}\left( \sum\limits_{p=1}^{P}{{{M}_{p}}} \right)$. The complexity of the proposed ITW relies on the number of iterations to achieve the optimal weighting factor, i.e., $\mathcal{O}\left( \sum\limits_{i=1}^{I}{P} \right)$. The computational complexity of the proposed fusionNet is associated with the number of neurons and the depth of the FCNN, so it is expressed as $\mathcal{O}\left( \sum\limits_{p=1}^{P}{{{L}_{e}}{{P}^{2}}{{M}_{p}}\prod\limits_{d=1}^{D}{{{L}_{d}}}} \right)$, where ${L}_{d}$ and ${L}_{e}$ are the number of neurons and the epochs, respectively.
The complexity of the total proposed methods is summarized in Table~\ref{complexity}.

\begin{table*}[!t]
\caption{The complexity of the proposed methods\label{complexity}} 
\centering
\begin{tabular}{|c||p{10cm}|}
\hline
Methods & Complexity \\ 
\hline
\multirow{2}{*}{Proposed ITW-GMin/GMaxCS} & 
$\mathcal{O}\left( M^{3} + 8M^{2} + ML(2M + 3) - 8N + 1 \right) +$ \\ 
& $\mathcal{O} \left( \sum\limits_{p=1}^{P} K_{p}N_{p} + 4(K_{p} - 1)(2K_{p} + N_{p}K_{p} - 2) \right) +$ \\
& $\mathcal{O} \left( \sum\limits_{p=1}^{P} M_{p} + \sum\limits_{i=1}^{I} P \right)$ \\
\hline
\multirow{2}{*}{Proposed fusionNet-GMinD/GMaxCS} & 
$\mathcal{O}\left( M^{3} + 8M^{2} + ML(2M + 3) - 8N + 1 \right) +$ \\ 
& $\mathcal{O} \left( \sum\limits_{p=1}^{P} K_{p}N_{p} + 4(K_{p} - 1)(2K_{p} + N_{p}K_{p} - 2) \right) +$ \\
& $\mathcal{O} \left( \sum\limits_{p=1}^{P} M_{p} + \sum\limits_{p=1}^{P} L_{e} P^{2} M_{p} \prod\limits_{d=1}^{D} L_{d} \right)$ \\
\hline
\end{tabular}
\end{table*}

\section{Simulation Results}\label{sec_simu}

In this section, simulation results are presented to validate the effectiveness of the proposed methods. 
The important parameter settings used in our simulations are provided in Table \ref{tableDOA}. Besides, the root-mean-squared error (RMSE) is used to evaluate the effect, expressed by

\begin{align} \label{}
	RMSE = \sqrt {\frac{1}{T}\sum_{t}^{T}(\hat{\theta}_{t}-\theta_0)^2}
\end{align}
where $T$ is the number of Monte Carlo experiments.

\begin{table}[!t]
\caption{System and simulation parameter settings}\label{tableDOA}
\centering
\begin{tabular}{|c||c|}
\hline
Parameters  & Values   \\ 
\hline
Number of groups: $P$ & 3 \\
\hline
Antennas in each group: $M_1, M_2, M_3$     & 7, 11, 13 \\
\hline
Subarrays in each group: $K_1, K_2, K_3$ & 16, 16, 16 \\
\hline
Antennas of FD array & 128  \\
\hline
Number of snapshots: $H$& 100 \\
\hline
Transmitter direction: $\theta_0$ & 41$^\circ$    \\
\hline
Monte Carlo experiments: T & 3000   \\
\hline
\end{tabular}
\end{table}

\begin{figure}[!htb]
	\centering	\includegraphics[width=4.0in]{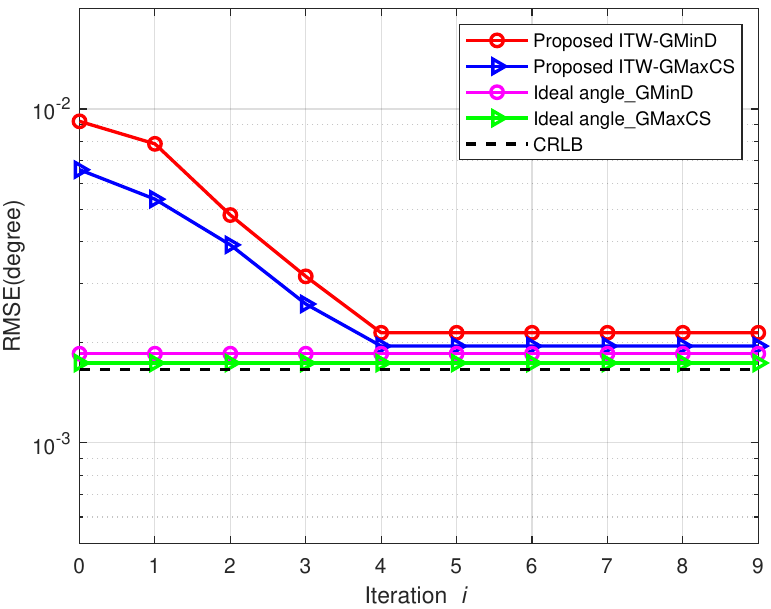}
	\caption{RMSE versus number of iterations curves.}\label{iteration curve}
\end{figure}

Figure \ref{iteration curve} shows the RMSE curves of our proposed ITW methods versus the number of iterations with a SNR$=10$dB, evaluating the convergence performance across $9$ iterations. Meanwhile, the performance obtained with the ideal angle and the CRLB are utilized as benchmarks. From Figure.\ref{iteration curve}, it can be observed that the RMSE of the proposed Co-ITW-GMaxCS and Co-ITW-GMinD methods converge within $4$ iterations. Furthermore, the convergence of Co-ITW-GMaxCS is faster and more stable compared to Co-ITW-GMinD. This is attributed to the fact that our proposed Co-ITW-GMaxCS better reflects the similarity between signals and more accurately captures their directional characteristics, resulting in more reasonable weight updates in each iteration.

\begin{figure}[!htb]
	\centering	\includegraphics[width=4.0in]{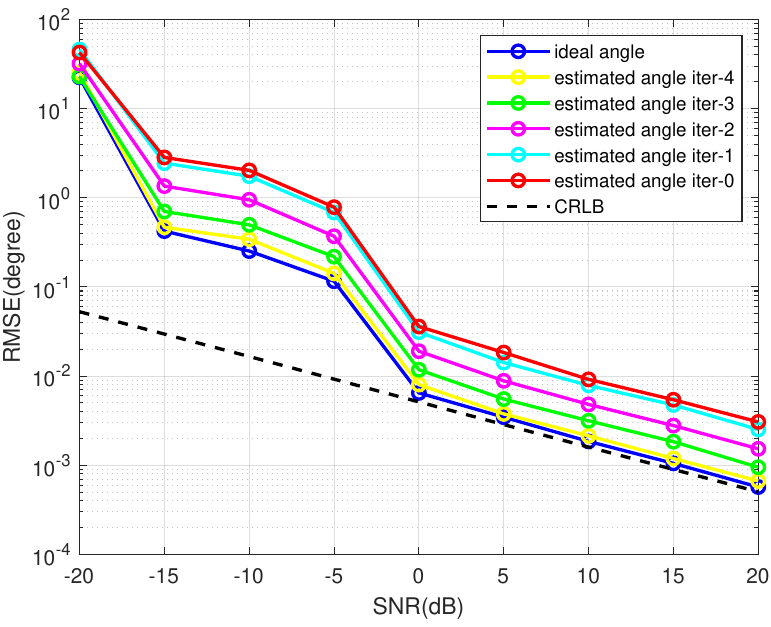}
	\caption{RMSE versus SNR of the proposed Co-ITW-GMinD method.}\label{ED_RMSE_SNR}
\end{figure}

\begin{figure}[!htb]
	\centering	\includegraphics[width=4.0in]{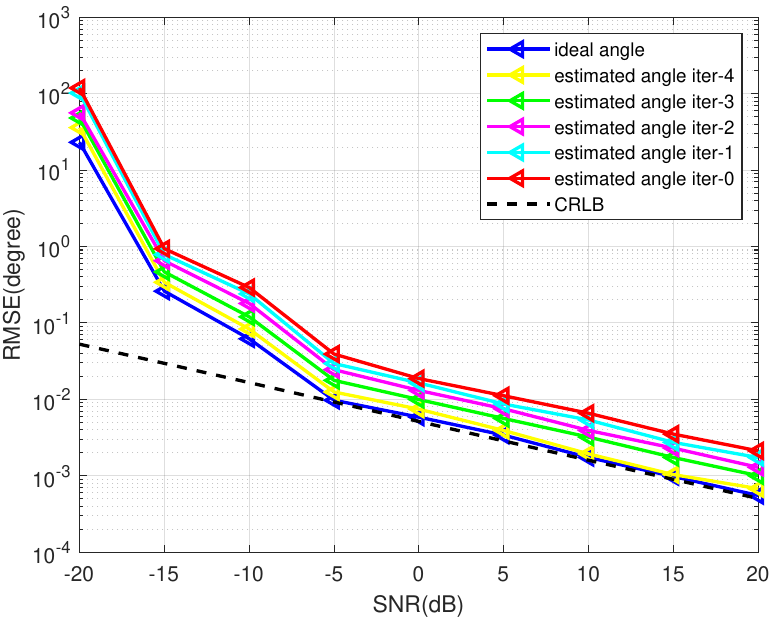}
	\caption{RMSE versus SNR of the proposed Co-ITW-
GMaxCS method.}\label{cos_RMSE_SNR}
\end{figure}

Figure \ref{ED_RMSE_SNR} and Figure \ref{cos_RMSE_SNR} plot the curves of RMSE versus SNR for the proposed Co-ITW-GMinD and Co-ITW-GMaxCS approaches, respectively. As shown in the figures, there is a approximately $3$ to $6$ times performance difference between the RMSE obtained by calculating weighting coefficients with estimated angles and those obtained with the ideal angles.
And the performance of both proposed methods gradually approaches the ideal performance through a certain number of iterations ($ \ge 4$). Noteworthy, the proposed Co-ITW-GMaxCS exhibits more excellent DOA performance compared to the Co-ITW-GMinD in extremely low SNR settings.

\begin{figure}[!htb]
	\centering	\includegraphics[width=4.0in]{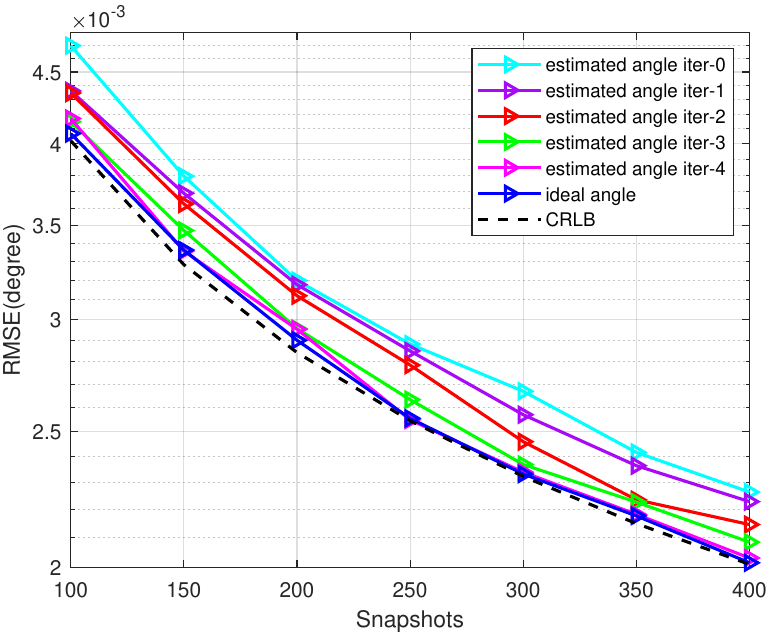}
	\caption{RMSE versus number of snapshots of the proposed Co-ITW-GMinD method.}\label{ED_snap_RMSE}
\end{figure}

\begin{figure}[!htb]
	\centering	\includegraphics[width=4.0in]{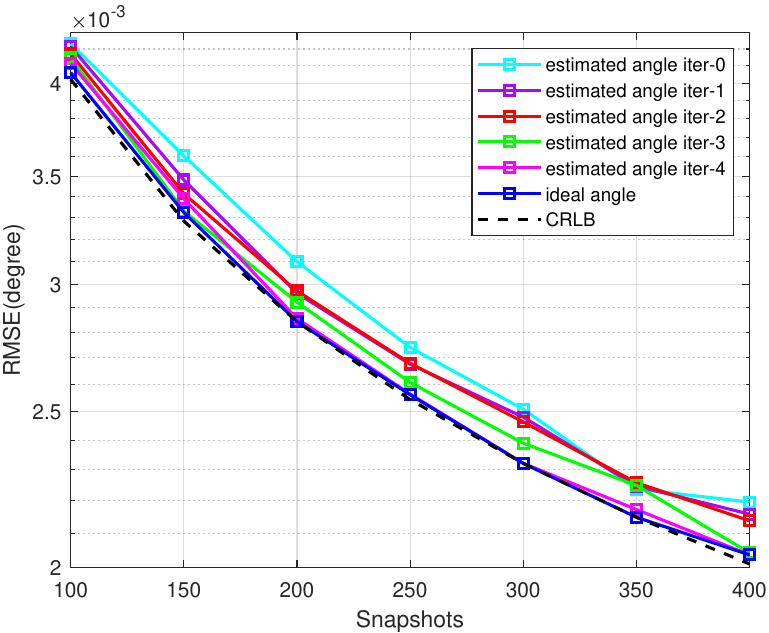}
	\caption{RMSE versus number of snapshots of the proposed Co-ITW-GMaxCS method.}\label{cos_RMSE_snap}
\end{figure}

Figure \ref{ED_snap_RMSE} and Figure \ref{cos_RMSE_snap} 
illustrate the RMSE curves for our proposed Co-ITW-GMinD and Co-ITW-GMaxCS methods versus the number of snapshots $H$ at SNR $= 0$dB, respectively. From these figures, 
RMSE performance improves progressively with an increasing number of snapshots $H$ and iterations. 
Consistent with the RMSE versus SNR results, the proposed Co-ITW-GMaxCS shows a more significant advantage over Co-ITW-GMinD at small-number snapshots scenarios ($H \le 250$). Combining insights from Figure \ref{ED_RMSE_SNR} and Figure \ref{cos_RMSE_SNR}, the proposed Co-ITW-GMaxCS exhibits superior DOA performance in both ultra-low snapshot and SNR regions.

\begin{figure}[!htb]
	\centering	\includegraphics[width=4.0in]{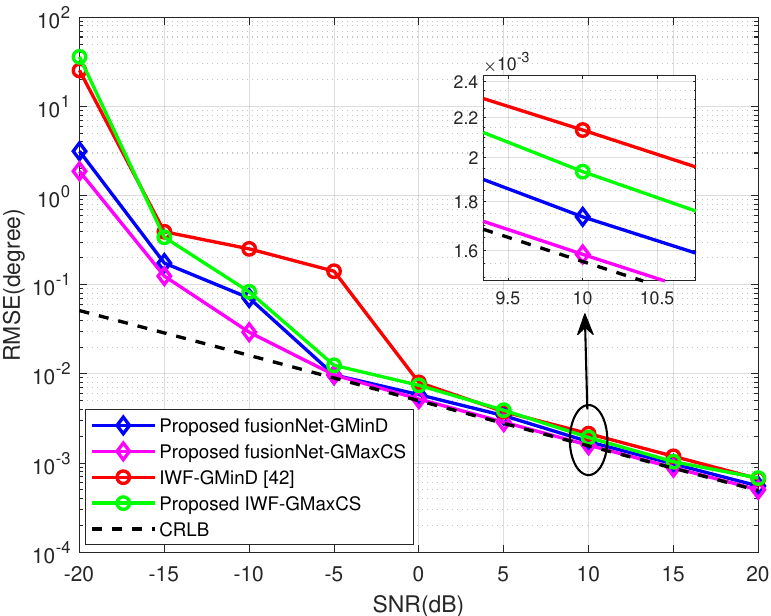}
	\caption{RMSE versus SNR of the proposed method.}\label{FUSION_ITW}
\end{figure}

Figure.\ref{FUSION_ITW} illustrates the RMSE versus SNR for our proposed four approaches and CRLB under $H=100$.
From Figure.\ref{FUSION_ITW}, the proposed methods can realize the CRLB when SNR $\ge 0$ dB. It is noteworthy that the proposed Co-fusionNet-GMinD and Co-fusionNet-GMaxCS exhibit superior DOA estimation performance, approaching the CRLB at SNR $= -5$dB. And yet our proposed Co-ITW-GMinD and Co-ITW-GMaxCS methods are more sensitive to high SNR situations.
Thus, the proposed fusionNet\_based methods are more outstanding than the Co-ITW\_based methods in the extreme-low SNR range. Particularly, the accuracy of the former is eight times that of the latter at SNR $= -20$dB.

\section{Conclusions}\label{sec_con}
In this paper, we proposed a new co-learning-assisted three-stage green fusion passive DOA framework for H$^2$AD MIMO receiver arrays, which effectively realizes high-precision, high time-efficiency, low-cost, and more practical DOA estimators. On this basis, an efficient clustering methods, called  GMinD, was designed to successfully eliminate the phase ambiguity and acquires a series of group-true solutions. Thereafter, two high-performance fusion methods, called IWF and fusionNet, were presented to integrate all subarray group-true angles and the rough estimated angles for the FD subarray, thereby obtaining the final DOA estimation value. 
More specifically, the IWF provides an efficient solution for estimating source directions in realistic communication scenarios where noise is always non-negligible. It also significantly mitigates the DOA performance loss caused by the direct use of estimated angles. The method's key characteristic lies in its capability to compensate for performance loss caused by noise through iterative operations. Furthermore, the fusionNet, designed as a two-layer FCNN, achieves higher-resolution DOA measurement. 
Experimental results showed that our proposed four approaches can achieve the CRLB and the desired DOA performance. Particularly, in the ultra-low SNR region (SNR $< 0$ dB), the proposed MM-fusionNet-GMinD and MM-fusionNet-GMaxCS show superior DOA measurement performances compared to the proposed MM-IWF-GMaxCS and MM-IWF-GMinD.
In conclusion, the proposed framework and approaches can quickly eliminate the phase ambiguity with higher time-efficiency, low cost and low complexity while maintaining the same low-latency as the FD MIMO structure. Therefore, it is promising for application in next-generation green wireless communication systems such as 6G.


\end{document}